\documentclass[letterpaper, 10 pt, conference]{ieeeconf}  

\IEEEoverridecommandlockouts

\usepackage{times}
\usepackage{multicol}
\usepackage[bookmarks=true]{hyperref}
\usepackage{tikz}
\usepackage{xcolor}
\usepackage{hyperref}
\usepackage{amssymb}
\usepackage{amsmath, amsfonts}
\usepackage{graphicx}
\usepackage{siunitx}
\usepackage{standalone}
\usepackage{booktabs}
\usepackage[ruled,vlined]{algorithm2e}
\usepackage{mdframed}
\usepackage{fancyvrb}
\usepackage{soul}
\usepackage{dsfont,mathabx}
\usepackage[font=footnotesize, skip=5pt]{caption}

\title{\LARGE \bf
Learning Forceful Manipulation Skills from Multi-modal \\ Human Demonstrations
}

\author{An T. Le$^{1}$, Meng Guo$^{2}$, Niels van Duijkeren$^{3}$, \\ Leonel Rozo$^{2}$, Robert Krug$^{3}$, Andras G. Kupcsik$^{2}$, Mathias B\"urger$^{2}$.  
\thanks{$^{1}$University of Stuttgart; $^{2}$Bosch Center for Artificial Intelligence (BCAI), Germany; 
$^{3}$Bosch Corporate Research, Germany. 
Corresponding author: Meng Guo. Contact: \texttt{Meng.Guo2@de.bosch.com}.}%
}

\begin{document}
  
\maketitle

\begin{abstract}
Learning from Demonstration (LfD) provides an intuitive and fast approach to program robotic manipulators. 
Task parameterized representations allow easy adaptation to new scenes and online observations. 
However, this approach has been limited to pose-only demonstrations and thus only skills with spatial and temporal features. 
In this work, we extend the LfD framework to address forceful manipulation skills, which are of great importance for industrial processes such as assembly. 
For such skills, multi-modal demonstrations including robot end-effector poses, force and torque readings, and operation scene are essential. 
Our objective is to reproduce such skills reliably according to the demonstrated pose and force profiles within different scenes. 
The proposed method combines our previous work on task-parameterized optimization and  attractor-based impedance control. 
The learned skill model consists of (i) the attractor model that unifies the pose and force features, and (ii) the stiffness model that optimizes the stiffness for different stages of the skill. 
Furthermore, an online execution algorithm is proposed to adapt the skill execution to real-time observations of robot poses, measured forces, and changed scenes. 
We validate this method rigorously on a 7-DoF robot arm over several steps of an E-bike motor assembly process, which require different types of forceful interaction such as insertion, sliding and twisting. 
\end{abstract}

\section{Introduction}\label{sec:intro}
\begin{figure}[t]
  \centering
  \includegraphics[width=0.99\linewidth]{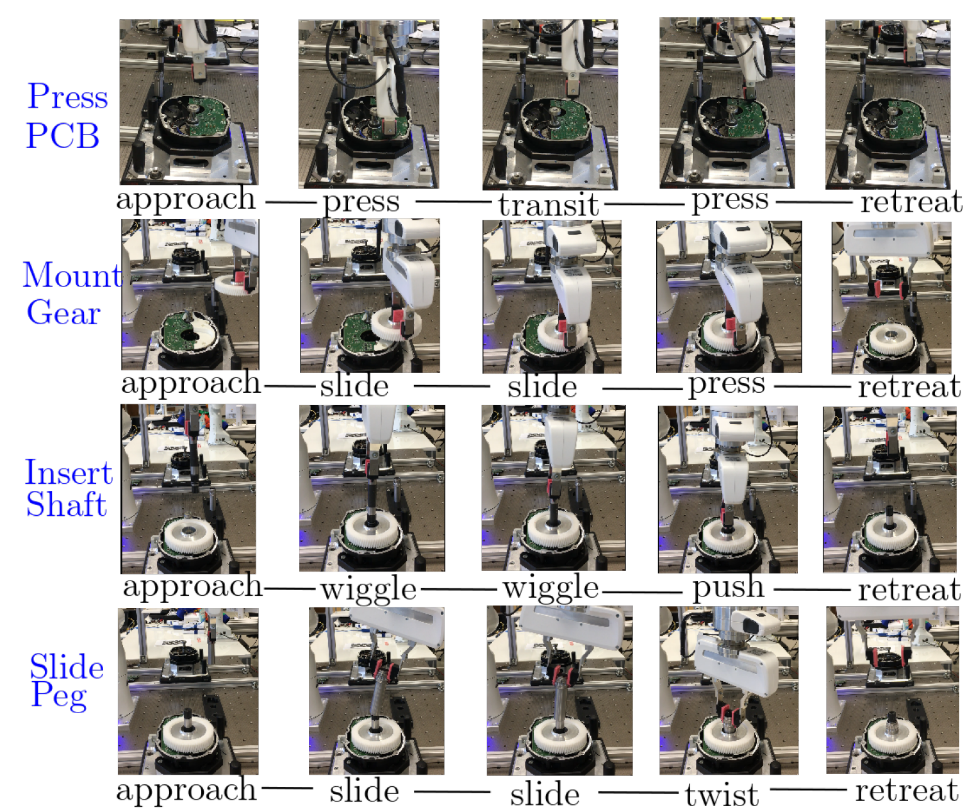}
  \caption{Various forceful skills considered in the experiment, each of which contains several stages of forceful interaction.}
  \label{fig:cases}
  \vspace{-0.4cm}
\end{figure}

Forceful interaction is vital for robotic manipulation in industry. 
While stiff kinematic trajectory tracking is adequate for simple pick-and-drop tasks, it is insufficient for tasks that involve explicit interaction with the environment. 
For instance, consider the E-bike motor assembly illustrated in Fig.~\ref{fig:cases}; which is also the use case in the experiments of Section \ref{sec:experiments}. 
After following an approaching free-space trajectory, a metallic shaft should be push firmly into a hole.
In contrast, a metallic peg ought to be slid over the metallic shaft pushing down softly while twisting to match the inline carvings of the peg and shaft.
These skills require significantly different kinematic trajectories, force trajectories and stiffness levels. 
Variable impedance control and learning have recently been exploited in manipulation scenarios characterized by the aforementioned conditions~\cite{Abudakka20survey}. 
Furthermore, compliant skills have shown to be useful in collaborative scenarios between robots and humans or between multiple robots~\cite{rozo2016learning, Amanhoud19:MotionForce_DynSyst},
  especially if the compliance is changed for different stages of the skill during interaction. 

Learning from demonstration (LfD) has increasingly shown to be an intuitive and effective way to program primitive skills for industrial robots, see~\cite{calinon2016tutorial, bogue2016europe}. 
Instead of simply re-playing the recorded demonstration, 
several parameterized skill models have been proposed to improve the generalization over different scenarios, see~\cite{calinon2016tutorial, niekum2015learning, Osa2018Imitation}.
However most of these approaches have focused on kinematic demonstrations, i.e., only the robot end-effector pose is recorded and used during learning. 
Such position-based models are often inadequate for tasks that require forceful interactions, 
as they not only neglect completely the demonstrated forces, but also rely on a manually-set stiffness. 

In this work, we exploit the LfD paradigm to address forceful manipulation skills.
More specifically, we consider multi-modal demonstrations including robot poses, force and torque readings, as well as registered scenes from visual perception. 
We propose a learning framework that combines our previous work on task parameterized optimization in~\cite{Schwenkel2019Optimizing, rozo2020learning} and attractor-based impedance learning in~\cite{rozo2016learning}. 
The learned skill model consists of two parts: \emph{(i)} a task-parametrized attractor model that unifies pose and force demonstration data; and \emph{(ii)} a stiffness model that specifies the optimal stiffness for different stages of the skill. 
Such combination allows the robot to reproduce the desired skill patterns extracted from the demonstrated pose and force profiles while adapting to different environment conditions. 

Last but not least, an import criteria of force-sensitive skills is the ability to adapt to real-time observations including deviations in robot poses due to tracking error, measured external forces, and changes in operation scenes. 
In this work, we propose an adaptation algorithm that optimally adapts the skill trajectory and the associated stiffness given such past observations.  

The remainder of the paper is organized as follows: 
Sec.~\ref{sec:related} reviews the related work. 
The problem formulation is formally given in Sec.~\ref{sec:problem}, 
whereas the proposed solution is described in Sec.~\ref{sec:solution}. 
Sec.~\ref{sec:experiments} presents experiment results for an E-bike assembly use case. 
Finally, Sec.~\ref{sec:conclusion} concludes with future directions of research.

\section{Related Work}\label{sec:related}

\subsection{Learning from Demonstration}\label{sec:lfd-review}
Compared with traditional motion planning~\cite{lavalle2006planning}, Learning from Demonstration (LfD) is an intuitive and effective way to transfer human skills to robots~\cite{calinon2016tutorial, Osa2018Imitation, Ravichandar20LfD}.
Teaching methods for LfD include kinesthetic teaching, tele-operation, and visual demonstrations~\cite{Ravichandar20LfD}. 
Various skill models are proposed to abstract these demonstrations, such as the full robot end-effector trajectory~\cite{Osa2018Imitation}, 
Dynamic Movement Primitives (DMPs)~\cite{Ijspeert13dmp},
Probabilistic Movement Primitives (ProMP)~\cite{paraschos2013probabilistic}, 
 or Task-parameterized Gaussian Mixture Models (TP-GMMs)~\cite{rozo2016learning, calinon2016tutorial, Zeestraten17riemannian}, which extend GMMs by incorporating observations from different perspectives so called task parameters, 
task-parametrized hidden semi-Markov models (TP-HSMMs)~\cite{Schwenkel2019Optimizing, rozo2020learning},
and deep neural networks~\cite{pathak2018zero, huang2019neural} that directly map observations to control inputs. 
In this work, we adopt TP-HSMMs as learning model, mainly for two reasons:
first, TP-HSMMs provide an elegant probabilistic representation of motion skills, which extracts temporal, sequential and spatial features from few human demonstrations. 
In contrast, TP-GMMs only encode spatial information;
second, task parameterization allows the model to generalize to new situations.
Furthermore, the aforementioned works do not consider multi-modal demonstrations including both pose and force data.
This work further extends the LfD framework to such data.

\subsection{Force-based and impedance learning}\label{sec:force_impedance}
Most works that utilize force readings in the learning process can be categorized according to the force control strategy: direct and indirect force control.
The former explicitly assumes a force feedback controller and thus a task-frame authority strategy is needed to select which Cartesian axes are position or force controlled~\cite{Kober15:ForcePrimivites, Conkey19hybridLfD,Khader20contactrich}. 
The latter exploits impedance control to indirectly control the forces required by the task~\cite{rozo2016learning,Amanhoud19:MotionForce_DynSyst}, which is the approach we leverage in this work.
Impedance controllers provide a compliant behavior in all phases of a contact task but are limited in their force tracking ability, mainly due to the incomplete knowledge about the environment. 
To cope with this limitation, two distinct methodologies are usually adopted: impedance and set-point adaptation. 
Impedance adaptation adjusts the controller parameters (e.g., inertia, damping, and stiffness) to improve tracking in response to force, position, or velocity measurements~\cite{Abudakka18varimp,Bogdanovic20contact}. 
Set-point adaptation improves force tracking by adjusting the controller set-point (e.g., the reference position) based on force tracking errors or on estimations of the environment's change in stiffness~\cite{rozo2016learning,Amanhoud19:MotionForce_DynSyst}. 
A learning framework for force-sensitive manipulation skills is proposed in~\cite{JohannsmeierGH19} that combines impedance control with parameter learning, which however requires \emph{manual} design of different stages of a skill, e.g., ``approach'', ``contact'', ``align'', and ``insert'' for the peg-in-hole skill. 
In this work we combine variable impedance, state-dependent adaptation of the dynamics attractor set-point, and TP-HSMM into a single LfD framework.
This allows a robotic manipulator to learn and reproduce contact-rich tasks that require different compliance levels, feature temporal patterns, and depend on task parameters related to objects of interest in the robot workspace. 
In contrast to the aforementioned works which tackle only a subset of these problems, our approach provides a suitable solution that addresses all these challenges, which naturally arise in complex industrial settings. 

\begin{figure*}[ht]
  \centering
  \includegraphics[width=0.98\textwidth]{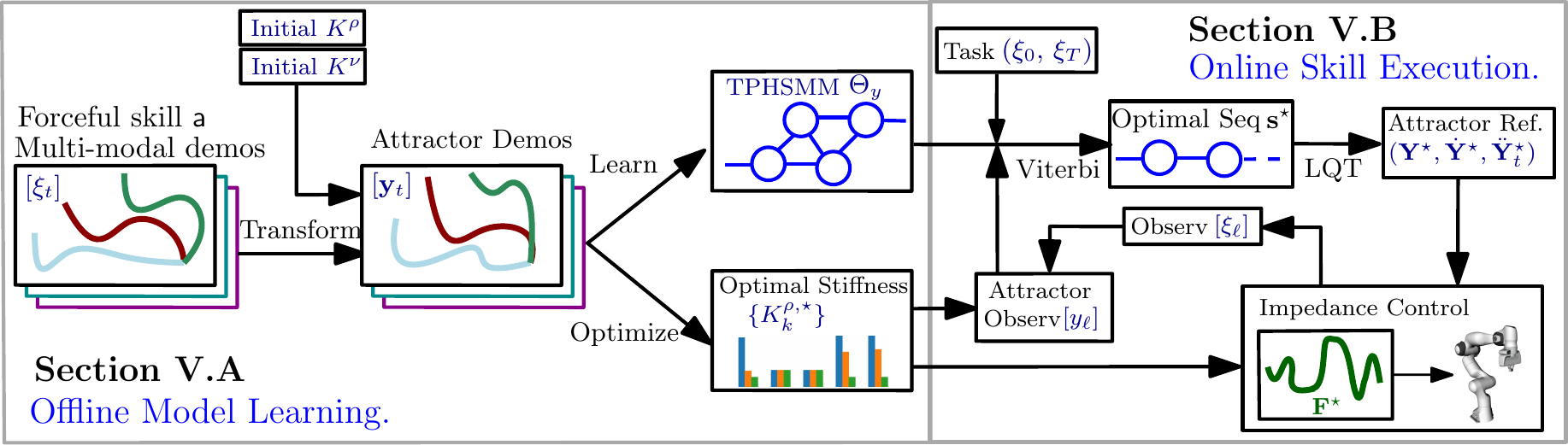}
  \caption{Overall diagram of the proposed method. 
Left: during the offline model learning, an attractor model $\mathbf{\Theta}_y$ and a stiffness model $\{\boldsymbol{K}_k^{\rho, \star}\}$ are learned from a set of multi-modal demonstrations; 
Right: during online skill execution, given a new scenario and a new task, the reference trajectory $\boldsymbol{Y}^\star$ and the preferred stiffness $\{\boldsymbol{K}_k^{\rho, \star}\}$ are generated online using the learned models and the real-time observations from the robot. 
They are then passed to the impedance controller to compute the actual control input $\boldsymbol{F}$.}
  \label{fig:diagram}
  \vspace{-0.4cm}
\end{figure*}
\section{Preliminaries}\label{sec:preliminary}
We briefly present some preliminary results in robot skill learning with focus on Task-Parameterized Hidden Semi-Markov Models (TP-HSMMs), 
as well as the algorithm to find the most-likely sequence of components within it.

\subsection{TP-HSMMs}\label{subsec:tp-hsmm}
As exploited in our earlier work~\cite{Schwenkel2019Optimizing, rozo2020learning}, TP-HSMM provides a \emph{compact} representation for both the temporal and spatial features of human demonstrations. 
Its task-parameterized formulation allows flexible adaption of robot skills to new scene conditions such as unseen object poses. 
 
Consider a set of demonstrations $\{\boldsymbol{\xi}\}=\{\big[\boldsymbol{\xi}_t\big]\}$,  where~$\boldsymbol{\xi}_t$ can belong to various manifolds, e.g., the Euclidean or the Riemannian manifold from~\cite{Zeestraten17riemannian}. 
Also, we assume that the same demonstrations are recorded from the perspective of $p=\{1,\cdots, P\}$ coordinate systems. 
These are given by the task parameters and include, e.g., objects of interest. 
One common way to obtain such data is to transform the demonstrations from a global frame to frame~$p$ by $\boldsymbol \xi_t^{(p)} = \boldsymbol A^{(p)^{-1}}(\boldsymbol \xi_t - \boldsymbol b^{(p)})$. 
Here, $\{(\boldsymbol b^{(p)},\boldsymbol A^{(p)})\}_{p=1}^P$ is the translation and rotation of frame $p$ w.r.t. the global frame. 
A Task-Parameterized HSMM (TP-HSMM) model is defined as: 
\begin{equation}\label{eq:tp-hsmm}
\mathbf{\Theta} = \left\{ \{a_{hk}\}_{h=1}^K,\, (\mu_k^D, \sigma_k^D),\, \gamma_k \right\}_{k=1}^K,
\end{equation}
where $\gamma_k=(\pi_k, \{(\boldsymbol \mu_k^{(p)},\, \boldsymbol\Sigma_k^{(p)})\}_{p=1}^P)$ is one component of the model, which is modeled as a task parameterized Gaussian mixture model (TP-GMM).
It represents the observation probability corresponding to component $k$; 
$a_{hk}$ is the transition probability from component~$h$ to~$k$; $(\mu_k^D,\, \sigma_k^D)$ describe the Gaussian distributions for the duration of component~$k$, i.e., the probability of staying in state $k$ for a certain number of consecutive steps.
Note that, differently from standard GMMs, the mixture model above cannot be learned independently for each frame. 
Indeed, the coefficients~$\pi_k$ are shared by all frames and the \mbox{$k$-th} component in frame~$p$ must map to the same \mbox{$k$-th} component in the global frame. 

The model parameters can be estimated using a reformulation of the Expectation-Maximization (EM) algorithm~\cite{calinon2016tutorial}, which is tailored to jointly train the HSMM and the underlying TP-GMM, with polynomial complexity. 
Once learned, the model $\mathbf{\Theta}$ can be used during skill reproduction to adapt automatically to new configurations of the $P$ task parameters.

\subsection{Most-Likely Sequence of Components}\label{sec:viterbi}
As motivated in~\cite{rozo2020learning, yu2003missingHSMM}, a common problem that arises with the HSMM model above is to find the \emph{most-likely} sequence of components within $\mathbf{\Theta}$, given the past observations $\big[\boldsymbol{\xi}_\ell\big]_{\ell=1}^t$ until time $t>0$ and the desired final observation $\boldsymbol{\xi}_T$.
This problem is relevant for robot motion generation, as we may need to estimate reference trajectories to achieve a specific task goal such as a desired pose of a manipulated object. 
More specifically, a modified Viterbi algorithm was proposed in our earlier work~\cite{rozo2020learning}, which defines: 
\begin{equation} \label{eq:viterbi}
\begin{aligned}
\delta_t(k) &= \max_{\forall d, \forall h\neq k}\left\{a_{hk} \delta_{t-d}(h)\, p_k(d)  \prod_{\ell=t-d+1}^t \tilde{b}_k(\boldsymbol \xi_{\ell})\right\},\\
\delta_1(k) &= b_k(\boldsymbol \xi_1)\,\pi_k\, p_k(1),
\end{aligned}
\end{equation}
where $p_k(d) = \mathcal{N}(d\,|\,\mu_k^D, \sigma_k^D)$ is the duration probability of component $k$, $\delta_t(k)$ is the likelihood of the system being in component $k$ at time $t$ \emph{and} not in state $k$ at $t+1$, see~\cite{forney1973viterbi} for details; and the observation probability at time $\ell$:
\begin{equation*}
\tilde{b}_k(\boldsymbol \xi_{\ell}) = \begin{cases}
\mathcal{N}(\boldsymbol \xi_{\ell}\,|\,\hat{\boldsymbol \mu}_k,\hat{\boldsymbol \Sigma}_k), &\, \ell \in\{1,2,...,t,\,T\};\\
1, &\, \ell \in \{t+1,t+2,...,T-1\}\,,
\end{cases}
\end{equation*}
where $(\hat{\boldsymbol \mu}_k,\hat{\boldsymbol \Sigma}_k)$ is the global Gaussian component~$k$ in $\mathbf{\Theta}$ given $\boldsymbol \xi_{\ell}$.
Namely, at each time $t$ and for each component~$k$, the two arguments that maximize equation $\delta_t(k)$ are recorded, and a simple backtracking procedure is used to find the most likely sequence of components, denoted by $\boldsymbol{s}^\star$.
This sequence optimally matches the given the observations and  the learned spatial-temporal distributions of the model.

\section{Problem description}\label{sec:problem}

Consider a multi-DoF robotic arm, whose end-effector has state $\boldsymbol{x} \in \mathbb{R}^3 \times S^3$ as its Cartesian position and orientation in the task space. 
However, for the sake of simplicity, the formulations in the sequel are all given for Euclidean space. 
Details regarding how quaternions should be handled properly are given in Sec.~\ref{subsec:manifold}. 
To achieve compliant behaviors, we assume that the arm is governed by a Cartesian Impedance Controller~\cite{hogan1985impedance} in the Lagrangian formulation:
\begin{equation} \label{eq:impedance}
  \boldsymbol{F} = \boldsymbol{K}^\rho (\boldsymbol{x}_d - \boldsymbol{x}) + \boldsymbol{K}^\nu (\dot{\boldsymbol{x}}_d - \dot{\boldsymbol{x}}) + \boldsymbol{I}(\boldsymbol q)\ddot{\boldsymbol{x}}_d + \Omega(\boldsymbol q, \dot{\boldsymbol q}), 
\end{equation}
where we omit the time $t$ as under-script for brevity; $\boldsymbol{F}$ is the input torque control, projected to task space;
$(\boldsymbol{x}_d, \dot{\boldsymbol{x}}_d, \ddot{\boldsymbol{x}}_d)$ are the desired pose, velocity and acceleration in the task space;
$\boldsymbol{K}^\rho$ and $\boldsymbol{K}^\nu$ are stiffness and damping terms; 
$\boldsymbol{I}(\boldsymbol{q})$ and $\Omega(\boldsymbol{q}, \dot{\boldsymbol{q}})$ are the task-space inertia matrix and internal dynamics terms, respectively; the latter terms depend on the current joint angular position $\boldsymbol{q}$ and velocity $\dot{\boldsymbol{q}}$, which are assumed to be available during execution. 

To demonstrate a forceful manipulation skill for an object, a human user performs several kinesthetic demonstrations on the robot for \emph{different} poses of the object. 
Particularly, the set of demonstrations is given by $\mathsf{D}=\{\mathsf{D}_1,\cdots, \mathsf{D}_{M}\}$, 
each of which is a \emph{timed} sequence of observations of the format:
\begin{equation}\label{eq:demos}
\mathsf{D}_m = \big[\boldsymbol{\xi}_t\big]_{t=1}^{T_m} = \left[\big{(}(\boldsymbol{x}_t, \dot{\boldsymbol{x}}_t, \ddot{\boldsymbol{x}}_t, \boldsymbol{f}_t), \boldsymbol{p}_{t} \big{)}\right]_{t=1}^{T_m}, 
\end{equation}
where at each time $t$ the observation $\boldsymbol{\xi}_t$ consists of the robot pose $\boldsymbol{x}_t$, velocity $\dot{\boldsymbol{x}}_t$, acceleration $\ddot{\boldsymbol{x}}_t$, the external force and torque $\boldsymbol{f}_t$, and finally the object pose $\boldsymbol{p}_t$. 
Such observations are often obtained from a state estimation module, a perception module or dedicated sensors.

The objective here is to learn a motion policy for the impedance controller in~\eqref{eq:impedance}, such that the skill can be reproduced reliably with the demonstrated pose and force profiles, even for new object poses.

Several examples of forceful skills addressed in the experiment section are shown in Fig.~\ref{fig:cases}. 
Note that these skills already show different characteristics of forceful manipulation, e.g., the PCB board requires a sequential pushing motion for pins, while for the gear a specific sliding path needs to be followed in a compliant way; 
the shaft should be inserted with a stiff downward motion, while for the peg a compliant twisting-and-sliding motion is needed.

\section{Proposed Solution}\label{sec:solution}
  
This section presents the two main components of the proposed solution, 
as shown in Fig.~\ref{fig:diagram}: 
the offline model learning as described in Sec.~\ref{subsec:learning} 
and the online skill execution as described in Sec.~\ref{subsec:execution}.

\subsection{Offline Model Learning} \label{subsec:learning}
Two models are learned offline from the set of demonstrations: 
(I) the attractor model~$\boldsymbol{\Theta}_y$, as a TP-HSMM model over the attractor trajectories;
 and (II) the stiffness model~$\{\boldsymbol{K}^{\rho, \star}_k\}$ associated with the attractor model.
Both models are essential as the attractor model shows how the pose trajectory should adapt to the actual scenario, while the stiffness model monitors the reproduction of the force profile.

\begin{figure}[t]
  \centering
  \includegraphics[width=0.45\textwidth]{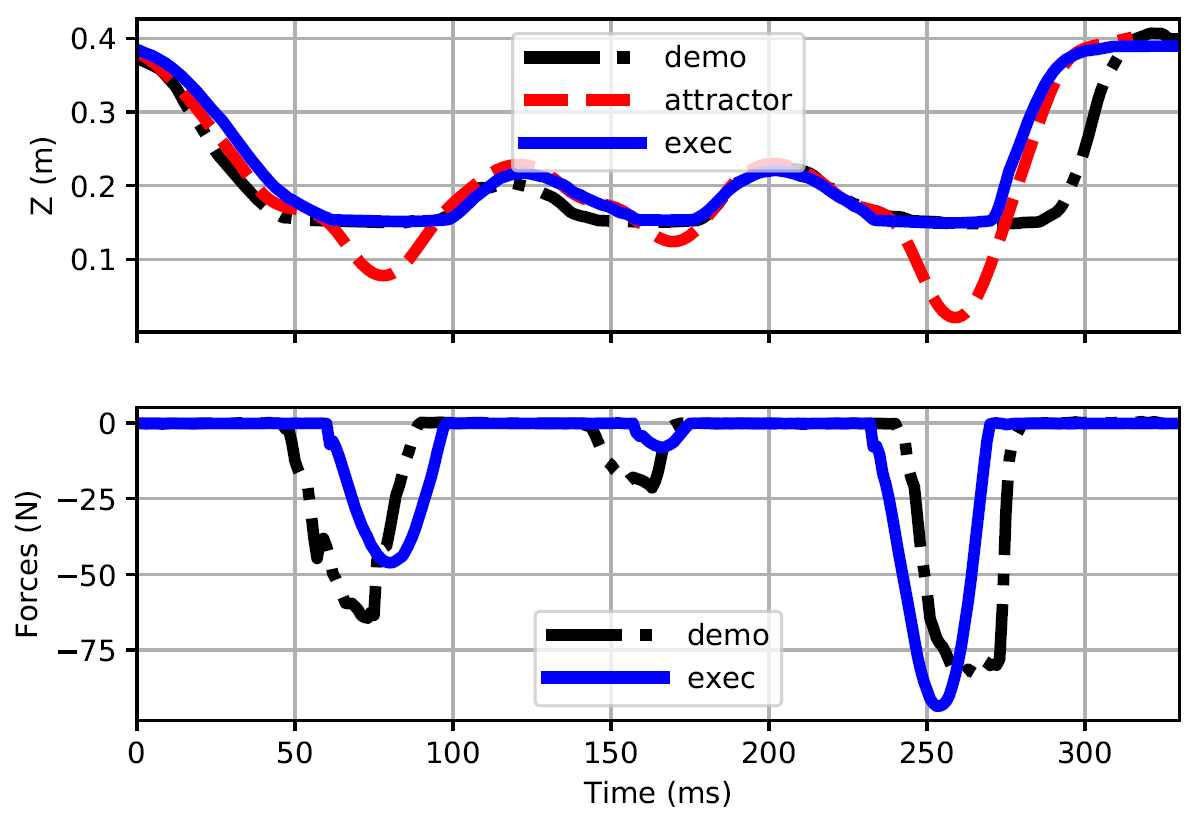}
  \caption{Top: trajectories in $z$-axis for the press-PCB skill: one demonstration, the learned attractor, and one execution.  
Bottom: the recorded force in $z$-axis during demonstration and one execution.}
  \label{fig:press-pcb-trajectory}
  \vspace{-0.4cm}
\end{figure}

\subsubsection{Learning of Attractor Model}\label{subsubsec:attractor}

One of the main challenges for the skill learning problem is to properly abstract an unified model from multi-modal demonstrations that encapsulate position, force and vision data.
We propose here to combine the attractor-based interaction model and the task-parameterized Markovian model for this purpose, both of which are developed in our earlier works~\cite{ rozo2016learning, Schwenkel2019Optimizing, rozo2020learning}.

In particular, we employ the attractor interaction model proposed in~\cite{rozo2016learning} to
transform the pose and force demonstrations to attractor trajectories by assuming that the attractor is driven by a virtual mass-spring-damper system. 
Consider any demonstration~$\mathsf{D}_m=\big[\boldsymbol{\xi}_t\big]$ from~\eqref{eq:demos}, the associated attractor trajectory $\big[\boldsymbol{y}_t\big]$ can be computed by:
\begin{equation} \label{eq:attractor}
\boldsymbol{y}_t = \boldsymbol{x}_t + \boldsymbol{K}^{-\rho}_t\left(\boldsymbol{K}^\nu_t\dot{\boldsymbol x}_t + \ddot{\boldsymbol x}_t - \boldsymbol f_t\right), 
\end{equation}
where $(\boldsymbol{x}_t, \dot{\boldsymbol{x}}_t, \ddot{\boldsymbol{x}}_t, \boldsymbol{f}_t)\in \boldsymbol{\xi}_t$ is part of the demonstration as described in Sec.~\ref{sec:problem}; 
$\boldsymbol{K}^\rho_t$, $\boldsymbol{K}^\nu_t$ are the stiffness and the damping terms, the design of which is described in the sequel;
$\boldsymbol{K}^{-\rho}_t=(\boldsymbol{K}^\rho_t)^{-1}$ for brevity. 
Intuitively, the position, velocity, acceleration and force demonstrations are transformed into a \emph{single} entity: the pose of a virtual attractor. 
Examples of the demonstrated pose trajectory and the computed attractor trajectory for the press-pcb skill are shown in Fig.~\ref{fig:press-pcb-trajectory}. 
It can be seen that the resulting attractor pose can differ greatly from the demonstrated pose when large velocities and sensed forces are present.

In other words,~\eqref{eq:attractor} allows us to transform each demo $\mathsf{D}_m \in \mathsf{D}$ into an attractor demo $\boldsymbol{\Psi}_m=\big[(\boldsymbol{y}_t,\, \boldsymbol{p}_t)\big]$, i.e., the attractor trajectory and the associated object pose. 
As a result, the standard procedure as described in Sec.~\ref{subsec:tp-hsmm} can be followed to learned a TP-HSMM model from the set of attractor demonstrations~$\boldsymbol{\Psi} = \{\boldsymbol{\Psi}_m\}$. 
First, the attractor trajectory $\big[\boldsymbol{y}_t\big]$ is transformed into local observations from different frames, e.g., from the initial robot pose and the object pose. 
Then, an EM algorithm is used to compute the TP-HSMM model as defined in~\eqref{eq:tp-hsmm}, which encapsulates the spatial-temporal features of the derived attractor trajectories. 
Denote by $\mathbf{\Theta}_y$ this attractor model, which can already adapt to different initial robot and object poses due to its task parameterization.
The attractor models associated with the last stage of pushing a shaft and twisting a peg are shown in Fig.~\ref{fig:stiffness-exp}. 
Typically, the Gaussian components have small covariance during contact between the robot and the workstation with large forces, while a larger variance is allowed during free motion.

\subsubsection{Optimization of Stiffness Matrix}\label{subsubsec:learn-stiffness}
As mentioned earlier, to compute the attractor trajectories in~\eqref{eq:attractor}, the stiffness and damping terms  $\boldsymbol{K}^\rho_t$ and $\boldsymbol{K}^\nu_t$ have to be chosen beforehand. 
Clearly, the choice of these terms has a great impact on the resulting attractor model~$\mathbf{\Theta}_y$. 
In this section, we describe how to optimize them.

\begin{figure}[t]
  \centering
  \includegraphics[width=0.48\textwidth]{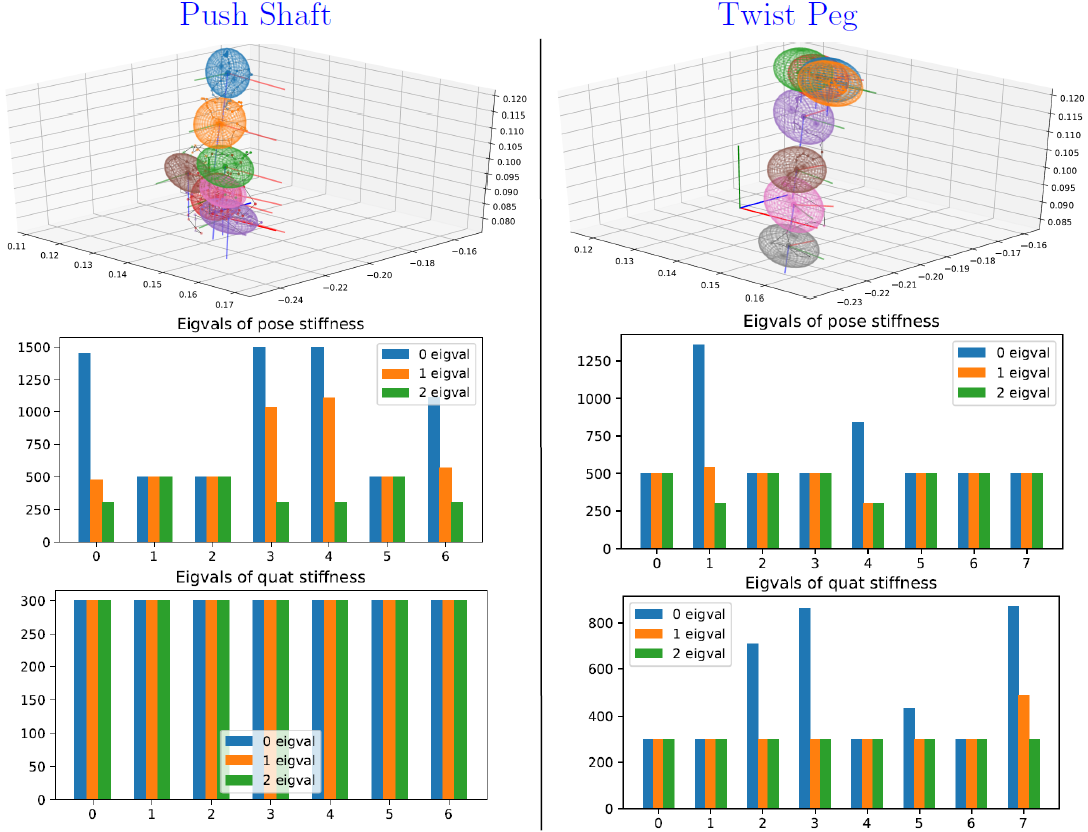}
  \caption{The learned attractor model (top), the optimized translation stiffness (middle) and the optimized angular stiffness (bottom), for the last stage of push-shaft skill (left) and twist peg (right).}
  \label{fig:stiffness-exp}
  \vspace{-0.4cm}
\end{figure}

Instead of solving them for each time instant, we propose to optimize these terms locally for each \emph{component} within~$\mathbf{\Theta}_y$. 
Particularly, consider component $k$ within $\mathbf{\Theta}_y$. 
For each attractor trajectory $\boldsymbol{\Psi}_m$, the accumulative residual of the computed attractor trajectory with respect to this component is given by:
\begin{equation}\label{eq:error}
\varepsilon_m = \sum_{\boldsymbol{\xi}_t \in \mathtt{D}_m} p_{t,k}\left(\boldsymbol{\mu}_k - \boldsymbol{x}_t - \boldsymbol{K}_k^{-\rho} \left(\boldsymbol{K}_t^{\nu}\dot{\boldsymbol{x}}_t + \ddot{\boldsymbol{x}}_t - \boldsymbol{f}_t \right)\right),
\end{equation}
where $p_{t,k}$ is the probability of state $\boldsymbol{x}_t$ belonging to component $k$, which is a by-product of the EM algorithm when deriving $\mathbf{\Theta}_y$; 
$\boldsymbol{\mu}_k$ is the mean of component $k$ from $\mathbf{\Theta}_y$; 
$(\boldsymbol{x}_t, \dot{\boldsymbol{x}}_t, \ddot{\boldsymbol{x}}_t, \boldsymbol{f}_t) \in \boldsymbol{\xi}_t$ is the demonstration point at time $t$ of $\mathsf{D}_m$; 
$\boldsymbol{K}_k^{-\rho}$ is the inverse of the stiffness term to be optimized, while the damping term~$\boldsymbol{K}_t^{\nu}$ remains unchanged. 
Consequently, the optimal local stiffness  for component $k$ can be computed by minimizing the complete residual over all demonstrations, namely:
\begin{equation}\label{eq:whole-error}
\boldsymbol{K}_k^{\rho,\star} = \underset{\boldsymbol{K}^\rho_{k}}{\textbf{min}}\; \left\|\sum_{\mathtt{D}_m}\, \varepsilon_m \right\|\, , \quad \text{s.t.}\; \boldsymbol{K}^\rho_{k} \succeq \boldsymbol{0},
\end{equation}
which requires the stiffness matrix to be positive semidefinite.
The above optimization problem belongs to the semidefinite program (SDP)~\cite{rozo2016learning}, which can be solved efficiently using techniques such as interior-point methods~\cite{ boyd_vandenberghe_2004}.

To summarize, an initial choice of $\boldsymbol{K}_t^\rho$ and $\boldsymbol{K}_t^\nu$ is set to compute the attractor model as described in~\ref{subsubsec:attractor}.
A common choice is the default stiffness of the underlying impedance controller in~\eqref{eq:impedance} and its critical damping term.
Afterwards, the local stiffness of each component can be optimized by~\eqref{eq:whole-error} as described above, denoted by $\{\boldsymbol{K}^{\rho, \star}_k\}$. 
Note that the learned stiffness varies along the attractor trajectory in order to match the robot stiffness during kinesthetic teaching. 
It becomes apparent in the sequel that the optimized stiffness is also crucial during online execution to react in real-time to pose and force/torque measurements. 
Fig.~\ref{fig:stiffness-exp} shows the optimized translational and rotational stiffness for the push-shaft and twist-peg skills. 
It can be seen that insertion requires relatively high translational stiffness upon contact while twisting requires high rotational stiffness.

\subsubsection{Riemannian Manifold Formulation}\label{subsec:manifold}
It is commonly the case in robotic manipulation that the end-effector pose contains orientation representations such as quaternions. 
Classical Euclidean-based methods often rely on local approximations, which imposes no guarantee on the validity of the results. 
As shown in~\cite{rozo2020learning, Zeestraten17riemannian}, the theory of Riemannian manifold can tackle this issue in an elegant way. 
More specifically, for each point $\boldsymbol{x}$ in the manifold $\mathcal{M}$, there exists a tangent space $\mathcal{T}_{\boldsymbol x} \mathcal{M}$. 
This allows us to carry out Euclidean operations locally, while being geometrically consistent with the manifold constraints.
Two special operations called exponential and logarithmic maps allows us to map points between $\mathcal{T}_{\boldsymbol x} \mathcal{M}$ and $\mathcal{M}$, while maintaining the geodesic distance. 
Another useful operation is the parallel transport, which moves vectors between tangent spaces without introducing distortion. 
The exact form of these operations for various manifolds can be found in~\cite{Zeestraten17riemannian}. 

The aforementioned calculations in this paper can be easily adapted to Riemannian manifold formulation. 
For instance, the subtraction of poses within~\eqref{eq:impedance} and~\eqref{eq:error} can be replaced by the logarithmic operation and the summation of poses in~\eqref{eq:attractor} by the exponential operation. 
More importantly, the TP-GMMs within the attractor model $\mathbf{\Theta}_y$  belong to the manifold. 
The Gaussian mixtures need to be computed \emph{iteratively} by projecting to the tangent space and back to the manifold. 
Thus the Riemannian formulation is often more computationally expensive than its Euclidean counterpart, which however always ensures the validity of the results.

\subsection{Online Skill Execution} \label{subsec:execution}
After both the attractor model and stiffness model are learned offline, they can be used for skill execution. 
The skill execution consists of two steps: initial synthesis and online adaption, which are described in this section.

\begin{figure}[t]
  \centering
  \includegraphics[width=0.8\linewidth]{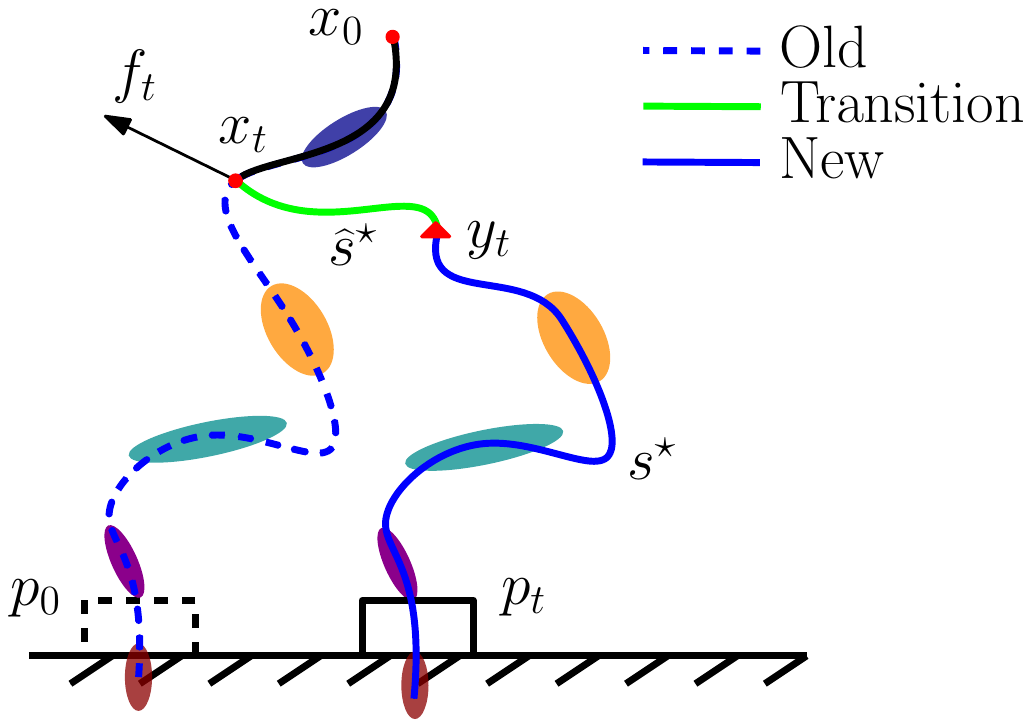}
  \caption{Online adaptation scheme during execution, due to changes in object pose $\boldsymbol{p}_t$, robot pose $\boldsymbol{x}_t$ and external forces $\boldsymbol{f}_t$. 
A transition phase (in green) is added before tracking the new attractor trajectory (in solid blue).}
  \label{fig:online-replanning}
  \vspace{-0.1cm}
\end{figure}

\subsubsection{Initial Synthesis}\label{subsubsec:offline-retrieval} %
Consider a new scenario where the robot and object poses may be different than the demonstrated ones. 
The first step is to compute the current frames for the attractor model $\mathbf{\Theta}_y$ given this new scenario, i.e., the $P$ frames in~\eqref{eq:tp-hsmm}. 
Second, the global GMMs in the global frame associated with $\mathbf{\Theta}_y$ are computed as weighted product of the local GMMs in these local frames. 
Moreover, given the initial observation $\boldsymbol{\xi}_0$ and possibly the desired final observation $\boldsymbol{\xi}_T$, the modified Viterbi algorithm in~\eqref{eq:viterbi} is used to compute the most-likely sequence of components within $\mathbf{\Theta}_y$, denoted by $\boldsymbol{s}^\star=[s_t^\star]$. 
Lastly, a linear quadratic tracking (LQT)~\cite{bemporad2002explicit} algorithm is used to retrieve the optimal and smooth reference trajectory that tracks this sequence of Gaussian components. 
This trajectory is the reference attractor trajectory to track by the robot, including a consistent velocity and acceleration profile, denoted by $\boldsymbol{Y}^\star=[\boldsymbol{y}_t^\star]$, $\dot{\boldsymbol{Y}}^\star=[\dot{\boldsymbol{y}}_t^\star]$, $\ddot{\boldsymbol{Y}}^\star=[\ddot{\boldsymbol{y}}_t^\star]$, respectively. 
Details on the controller design and its adaptation to Riemannian manifold can be found in our earlier work~\cite{rozo2020learning}. 

Given $s_t^\star$, $\boldsymbol{y}_t^\star$, $\dot{\boldsymbol{y}}_t^\star$, $\ddot{\boldsymbol{y}}_t^\star$ at time $t>0$ above, the control input for the Cartesian Impedance controller in~\eqref{eq:impedance} can be computed as the torque input projected to task space: 
\begin{equation} \label{eq:control}
  \boldsymbol{F}_t = \boldsymbol{K}_{s_t^\star}^{\rho,\star} (\boldsymbol{y}^\star_t - \boldsymbol{x}_t) + \boldsymbol{K}_t^\nu (\dot{\boldsymbol{y}}_t^\star - \dot{\boldsymbol{x}}_t) + \boldsymbol{I}(\boldsymbol{q}_t)\ddot{\boldsymbol{y}}_t^\star + \Omega(\boldsymbol{q}_t, \dot{\boldsymbol{q}}_t), 
\end{equation}
where the stiffness $\boldsymbol{K}_{s_t^\star}^{\rho,\star}$ is the \emph{optimal} stiffness computed from~\eqref{eq:whole-error} that is associated with component~$s_t^\star$; 
$\boldsymbol{K}_t^\nu$ is the damping term according to the choice in~\eqref{eq:attractor};
$(\boldsymbol{x}_t, \dot{\boldsymbol{x}}_t)$ are the current robot end-effector pose and velocity;
$(\boldsymbol{q}_t, \dot{\boldsymbol{q}}_t)$ are the current joint angular position and velocity. 
Via this impedance controller, the robot tracks the desired attractor trajectory $\boldsymbol{Y}^\star$ with the desired stiffness.

\subsubsection{Online Retrieval} \label{subsubsec:online-retrieval}
Once the robot starts moving, observations such as current robot pose and force/torque readings are obtained, which can indicate deviations in skill execution due to for instance external disturbances or tracking errors. 
Furthermore, changes in the scene such as new object poses are also registered. 
This section addresses how to adapt the reference attractor trajectory and the associated stiffness given these real-time measurements. 

To begin with, the changes in object poses lead to changes of the task parameters in the attractor model $\mathbf{\Theta}_y$. 
Thus, the global GMMs associated with all components are updated by re-computing the product of local GMMs similar to the initial synthesis. 
Consequently, the observation probability within~\eqref{eq:viterbi} is changed and so is the most-likely sequence~$\boldsymbol{s}^\star$. 
More importantly, the past observations in~\eqref{eq:viterbi} is not empty anymore as in the initial synthesis. 
In particular, given the past observations of robot pose and force readings $[\boldsymbol{\xi}_\ell]=[(\boldsymbol{x}_\ell,\,\boldsymbol{f}_\ell)]$ until time $t$, their corresponding \emph{virtual} observations of attractor~$[\boldsymbol{y}_\ell]$ is given by~\eqref{eq:attractor}, 
where the stiffness and damping terms are set to the same as used during execution in the impedance controller~\eqref{eq:control}. 
Then, these converted observations of attractors are used to evaluate the updated emission probability of the whole sequence, i.e., 
\begin{equation*}
\tilde{b}_k(\boldsymbol{\xi}_{\ell}) = \begin{cases}
\mathcal{N}(\boldsymbol{y}_{\ell}\,|\,\hat{\boldsymbol \mu}_{s_\ell^\star},\hat{\boldsymbol \Sigma}_{s_\ell^\star}), &\, \ell \in\{1,2,...,t,\,T\};\\
1, &\, \ell \in \{t+1,t+2,...,T-1\}\,,
\end{cases}
\end{equation*} 
where $\boldsymbol{y}_\ell=\boldsymbol{x}_\ell + \boldsymbol{K}^{-\rho}_{s_\ell^\star}\left(\boldsymbol{K}^\nu_\ell\dot{\boldsymbol x}_\ell + \ddot{\boldsymbol{x}}_\ell - \boldsymbol{f}_\ell\right)$ is the observation of the virtual attractor. 
Lastly, this emission probability is used in the modified Viterbi algorithm~\eqref{eq:viterbi} to compute the \emph{new} optimal sequence of components $\boldsymbol{s}^\star$. 

\setlength{\textfloatsep}{5pt}
\begin{algorithm}[t]
  \caption{Learn Forceful Skills from Multi-modal Demonstrations} \label{alg:overall}
  \LinesNumbered
  \KwIn{$\mathsf{D}=\{\mathsf{D}_m\}$, $[\boldsymbol{\xi}_\ell]_{\ell=0}^t$ at time $t\leq 0$}
  \KwOut{$\boldsymbol{\Theta}_y$, $\{\boldsymbol{K}^{\rho, \star}_k\}$, $\boldsymbol{F}_t$}
  \tcc{Offline Learning, Sec.~\ref{subsec:learning}}
  Convert $\mathsf{D}_m$ to $\boldsymbol{\Psi}_m$\;
  Learn attractor model $\boldsymbol{\Theta}_y$\;
  Learn stiffness model $\{\boldsymbol{K}^{\rho, \star}_k\}$\;
  \tcc{Online Execution, Sec.~\ref{subsec:execution}}
  Compute $\boldsymbol{s}^\star$ and $\boldsymbol{Y}^\star$ given $\boldsymbol{\Theta}_y$ and $(\boldsymbol{\xi}_0, \, \boldsymbol{\xi}_T)$\;
  \While{Goal $\boldsymbol{\xi}_{T}$ is not reached and $t<T$}
  {
    Compute $\boldsymbol{F}_t$ given $\boldsymbol{y}_t^\star$ and $\boldsymbol{K}_t^{\rho,\star}$ by~\eqref{eq:control}\;
    Send $\boldsymbol{F}_t$ to impedance controller\;
    Obtain new observation $\boldsymbol{\xi}_t$\;
    Update $\boldsymbol{\Theta}_y$ given new $P$ frames\;
    Update $\boldsymbol{s}^\star$ given $[\boldsymbol{\xi}_\ell]$ and $\boldsymbol{\Theta}_y$\;
    Compute $\widehat{\boldsymbol{s}}^\star$ by~\eqref{eq:prepend}\;
    Update $\boldsymbol{y}_t^\star$ via LQT given $\widehat{\boldsymbol{s}}^\star$ and $\boldsymbol{\Theta}_y$\;
  }
\end{algorithm}

Given the updated sequence $\boldsymbol{s}^\star$, a \emph{transition phase} should be followed to drive the system from its {current} pose $\boldsymbol{x}_t$ to the associated attractor pose $\boldsymbol{y}_t$ for $t>0$.
This is critical during online adaptation as the attractor pose $\boldsymbol{y}_t$ and the robot pose $\boldsymbol{x}_t$ in~\eqref{eq:attractor} can be significantly different due to e.g., high velocity and acceleration, or large external force, while this difference is negligible when the system starts initially at $t=0$. 
As a result, the adapted trajectory~$\boldsymbol{Y}^\star$ should start from the current pose $\boldsymbol{x}_t$, cross the via-point $\boldsymbol{y}_t$, and then track the updated sequence of components as specified in $\boldsymbol{s}^\star$. 
To achieve this, an artificial Gaussian global component $k_y$ is introduced which has its mean at $\boldsymbol{y}_t$ and the same covariance as the first component in $\boldsymbol{s}^\star$, and the current stiffness as the desired stiffness in $\boldsymbol{K}_t^{\rho,\star}$. 
Moreover, this component is assigned a duration $d_y$, which is proportional to the distance between $\boldsymbol{x}_t$ and $\boldsymbol{y}_t$. 
Finally, this new sequence of component $k_y$ with $d_y$ is pre-pended to the updated sequence $\boldsymbol{s}^\star$, namely, 
\begin{equation}\label{eq:prepend}
\widehat{\boldsymbol{s}}^\star = (k_y \cdots k_y)\, \boldsymbol{s}^\star,
\end{equation}
where $k_y$ is repeated $d_y$ times. 
This updated sequence $\hat{\boldsymbol{s}}^\star$ can be tracked by the updated reference trajectory $\boldsymbol{Y}^\star$ computed via LQT. 
Consequently, this new reference trajectory is sent to the impedance controller as shown in~\eqref{eq:control} with the associated stiffness.  
Fig.~\ref{fig:online-replanning} illustrates the above adaptation process, where the reacted attractor guides the end-effector to compensate for disturbances in robot poses or  external forces as well as changes in the scene.

\subsubsection{Overall Algorithm} \label{subsubsec:overall}

The overall framework is summarized in Alg.~\ref{alg:overall}. 
Note that the offline learning process between Line 1-3 only needs to be done once. 
The learned attractor model $\boldsymbol{\Theta}_y$ and stiffness model $\{\boldsymbol{K}^{\rho, \star}_k\}$ are saved and loaded directly for each execution. 
During online execution, the past observations $[\boldsymbol{\xi}_\ell]$ are saved and used whenever an online adaptation is triggered.

\begin{figure}[t]
  \centering
  \includegraphics[width=0.95\linewidth]{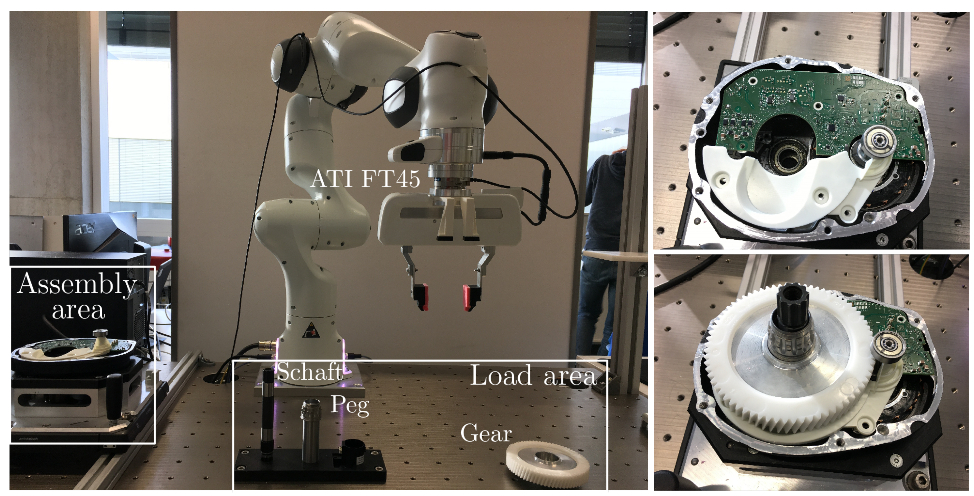}
  \caption{Left: workspace in experiment, including the assembly area and the loading area. Right: close-up of the E-bike motor internal.}
  \label{fig:workspace}
  \vspace{-0.1cm}
\end{figure}

\section{Experiments} \label{sec:experiments}
This section presents the experimental validation on a 7-DoF Franka Emika robot arm for an industrial E-bike motor assembly process. 
The arm is extended by a \emph{wrist-mounted} ATI FT45 sensor and a parallel gripper. 
The proposed approach is implemented in Python3. 
The Robot Operating System (ROS) enables communication between the planning, motion control and perception modules. 
All benchmarks are run on a desktop with an 8-core Intel Xeon CPU. 
Experiment videos can be found in the supplementary files. 

\subsection{Workspace Setup and Manipulation Tasks}\label{subsec:exp-setup}
As shown in Fig.~\ref{fig:workspace}, the workstation consists of the loading area where components are picked and the assembly area where the components are assembled together. 
The whole motor is roughly $16.5 \times 14.6 \times 15\si{\cm}^3$ in size. 
A pre-defined sequence of tasks should be followed during the assembly. 
As illustrated in Fig.~\ref{fig:cases}, we focus on the following four tasks as part of the assembly process in this experiment:

[Press-PCB]: 
A PCB of size $15.6 \times 7.5 \times 0.2 \si{\cm}^3$ is pressed into three pins on the motor base to secure the board. 
This skill consists four stages: approaching, pressing, transition, pressing and retreating. 
The pressing should be done with an appropriate force, if too small then the pins are not securely inserted; if too large then the PCB may be damaged; 
while the transition should be accurate to not miss the pins (with less than $\SI{2}{\mm}$ tolerance).  

\begin{table}[t]
\begin{adjustbox}{height=0.18\linewidth}
\begin{tabular}{ccccccc}
\toprule
Skill Name & $M$ & $T[\si{\second}]$  & $N$ &  $\mathsf{TP}$ & $t\, (\boldsymbol{\Theta}_y\,\vert \, \boldsymbol{K}^\star)$ [\si{\second}]\\ \midrule
\texttt{grasp\_gear} & $3$ & $1.2$ & $8$ &  $\{\mathfrak{r},\mathfrak{o}\}$ & $3\, \vert\,  2 $\\
\texttt{mount\_gear} & $3$ & $4.8$ & $18$ & $\{\mathfrak{r}, \mathfrak{g}\}$  & $14\, \vert\,  12 $\\
\texttt{pick\_shaft} & $3$ & $1.3$ & $7$ &  $\{\mathfrak{r}, \mathfrak{o}\}$ & $4\, \vert\,  2 $\\
\texttt{insert\_shaft} & $2$ & $5.7$ & $23$ &  $\{\mathfrak{r},\mathfrak{g}\}$ & $20\, \vert\,  18 $ \\
\texttt{pick\_peg} & $3$ & $2$ & $10$ &  $\{\mathfrak{r}, \mathfrak{g}\}$ & $3\, \vert\,  2 $ \\ 
\texttt{slide\_peg} & $2$ & $5.3$ & $24$ &  $\{\mathfrak{r}, \mathfrak{o}\}$  & $18\, \vert\,  15 $ \\
\texttt{press\_pcb} & $4$ & $6$ & $22$ & $\{\mathfrak{r}, \mathfrak{g}\}$  & $15\, \vert\,  10 $ \\
\bottomrule
\end{tabular}
\end{adjustbox}
\caption{For each skill, the number of demonstrations $M$, the trajectory length $T$, number of components $N$, choice of task parameters $\mathsf{TP}$, and the training time for $\boldsymbol{\Theta}_y$ and $\boldsymbol{K}^\star$. 
Note that $\mathfrak{r},\mathfrak{g}, \mathfrak{o}$ are the robot, global, object frame,  respectively. }
\label{table:skills}
\end{table}

[Mount-Gear]: 
A spur gear of size $11.5\times 11.5 \times 3.7\si{\cm}^3$ is mounted above the rotor casing. 
This skill consists of three stages: approaching, sliding and pushing. 
The sliding stage is quite delicate as the metal bottom of the gear needs to follow a tunnel on the casing into the desired location, followed by a light push to secure it in-place. 

[Insert-Shaft]: 
A drive shaft of size $1.6\times 1.6 \times 15\si{\cm}^3$ is inserted through the opening of the gear into a hole in the metal casing. 
This skill consists of three stages: approaching, wiggling and pushing. 
The ``wiggling'' stage is to insert the bottom of the shaft into the hole beneath the gear, which has a tolerance of around $1 \si{\milli\meter}$. 
Furthermore, the shaft is pushed with a \emph{large} force in a \emph{stiff} manner to ``click'' in position during the ``pushing'' stage. 
This proper placement is vital for the functionality. 

[Slide-Peg]:
The peg of size $2.4\times 2.4 \times 10.2 \si{\cm}^3$ is slided \emph{between} the inserted shaft and the mounted gear. 
This skill consists of four stages: attaching, sliding, twisting and pushing. 
The ``attaching'' stage is to attach the peg bottom to the top of the shaft, which has a toleration of $1 \si{\milli\meter}$. 
Moreover, to match the inline tooth of the shaft and the peg, the peg should be pushed with a \emph{small} force, while twisting in a \emph{compliant} manner during the this stage. 

It can be seen that these four skills already have different characteristics of forceful interaction. 
For instance, different stiffness or accuracy is required for different stages of the execution, of which translational and rotational stiffness can also be different.

\subsection{Results}\label{subsec:exp-results}
This section first presents the learning results of each skill above, and then the performance during reproduction in terms of success rate. 

\subsubsection{Skill Learning}\label{subsubsec:exp-learn}
Due to the high requirement on precision, all demonstrations are logged at $100 \si{\hertz}$.  
Details about the skill model such as the number of demonstrations, frames, skill duration are shown in Table~\ref{table:skills}. 
The initial stiffness and damping terms are set to $400$ and $40$ times identity matrix with appropriate dimensions.  
On average, it takes around $10\si{\second}$ to learn the skill model, where the EM for the attractor-model and the stiffness optimization for the stiffness model split the time equally. 
Examples of the learned attractor model and the associated stiffness model for the Insert-Shaft and Slide-Peg skills are shown in Fig.~\ref{fig:stiffness-exp}.  
Note that the optimized stiffness matches the expected behavior well, e.g., a high translational stiffness for insertion and a high rotational stiffness for twisting. 

\subsubsection{Skill Reproduction}\label{subsubsec:exp-reproduce}
As summarized in Alg.~\ref{alg:overall}, the learned model is applied on-line provided with the observed robot positions and external forces. 
Table~\ref{table:results} shows the success rate of executing each skill with $10$ repetitions.
The Press-PCB and Mount-Gear skills are reliably reproduced without any manual tuning of the associated frames, while a manual shift in the object frame is needed for the Insert-shaft and Slide-Peg skills to compensate for the tracking error of the underlying impedance controller. 
Such errors are mostly due to robot model mismatch during identification, and the internal joint force or torque control mechanism. 
Fig.~\ref{fig:press-pcb-trajectory} highlights the differences between the executed trajectory, the reference attractor trajectory and the demonstrations. 
In addition, it also shows that the exerted force profile during execution matches the demonstrations well.

\begin{table}[t]
\begin{center}
\begin{adjustbox}{width=0.95\linewidth}
\begin{tabular}{ccccc}
\toprule
Methods & Press-PCB & Mount-Gear & Insert-Shaft & Slide-Peg \\
\midrule
Pose-based & 0 & 3 & 0 & 1 \\
Demo-replay & 0 & 4 & 0 & 2 \\
Ours & 9 & 10 & 8 & 9 \\
Manual & N/A & N/A & 9 & 9 \\
\bottomrule
\end{tabular}
\end{adjustbox}
\caption{Success rate of four different methods out of $10$ repeated executions. 
Note that manual skills are not programmed for first two skills. }
\label{table:results}
\end{center}
\vspace{-0.35cm}
\end{table}

\begin{figure}[t]
  \centering
  \includegraphics[width=0.98\linewidth]{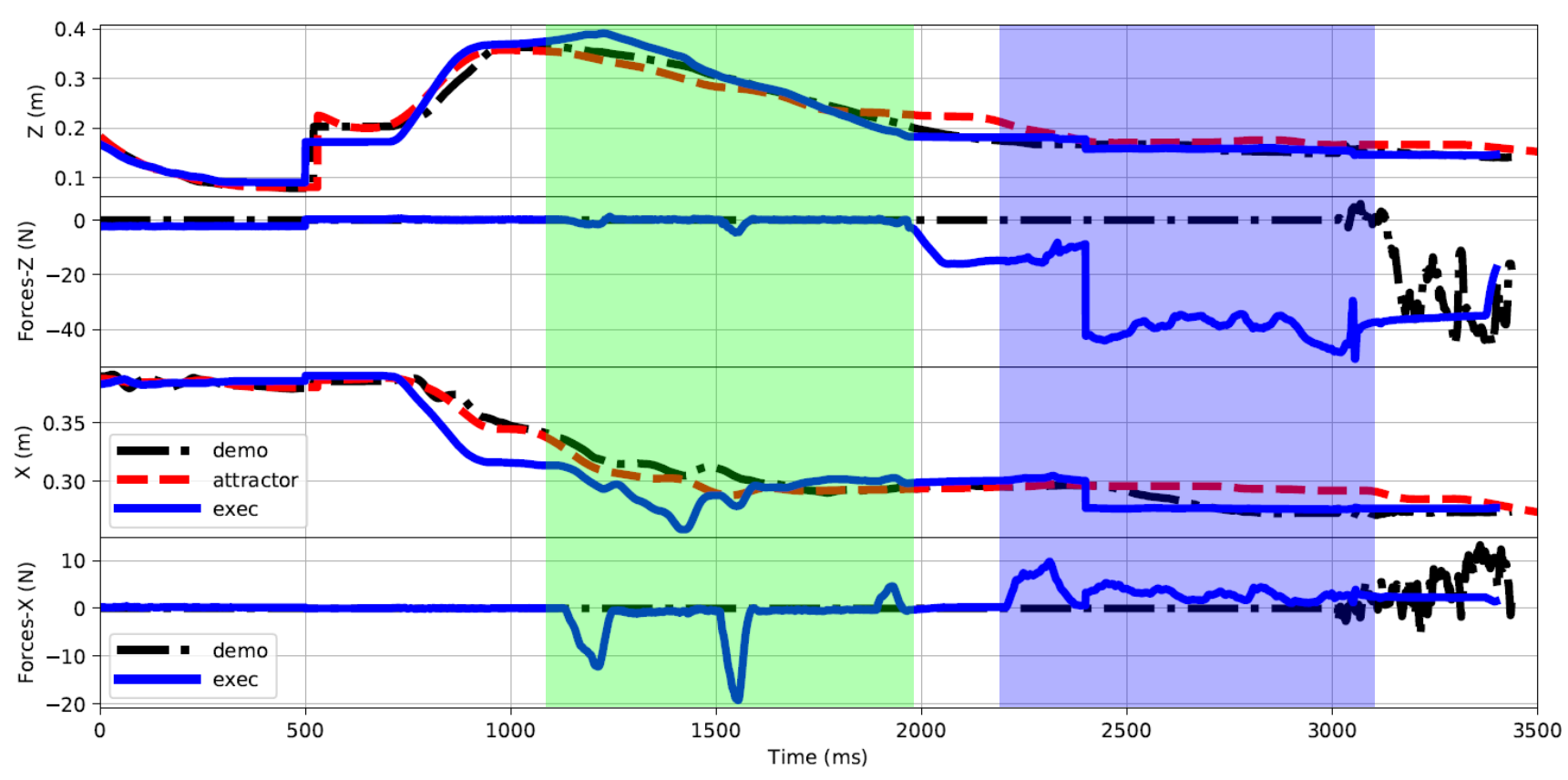}
  \caption{Online adaptation during the execution of Insert-Shaft skill. 
Position disturbance in robot $x$-axis is introduced within the green area, while force disturbance in the $z$-axis is added during insertion in the blue area.}
  \label{fig:online}
  \vspace{-0.1cm}
\end{figure}
%

To validate the online adaptation scheme, manual disturbances are introduced in robot position and measured forces during the execution of Insert-Shaft. 
As shown in Fig.~\ref{fig:online}, the execution adapts to these changes: 
the executed trajectory recovers to the original reference after the robot is pushed away along $x$-axis; 
and the exerted pushing force in $z$-axis remains close to the desired level after the robot wrist is lifted up or pushed down.

\subsection{Comparison and Discussion}\label{subsec:discussion}
For benchmark, the proposed method is compared against three main baselines:
the direct replay of the demonstration (demo-replay); 
the standard position-based skill model (pose-based) as proposed in~\cite{rozo2020learning}; 
and manually-tuned skills (manual) by following the procedure proposed in~\cite{JohannsmeierGH19}. 
Pose-based methods simply ignore the force profile when learning the skill model, while 
the manual method require manual tunning of \emph{all} fixed reference trajectory and the associated stiffness. 

\begin{figure}[t]
  \centering
  \includegraphics[width=0.95\linewidth]{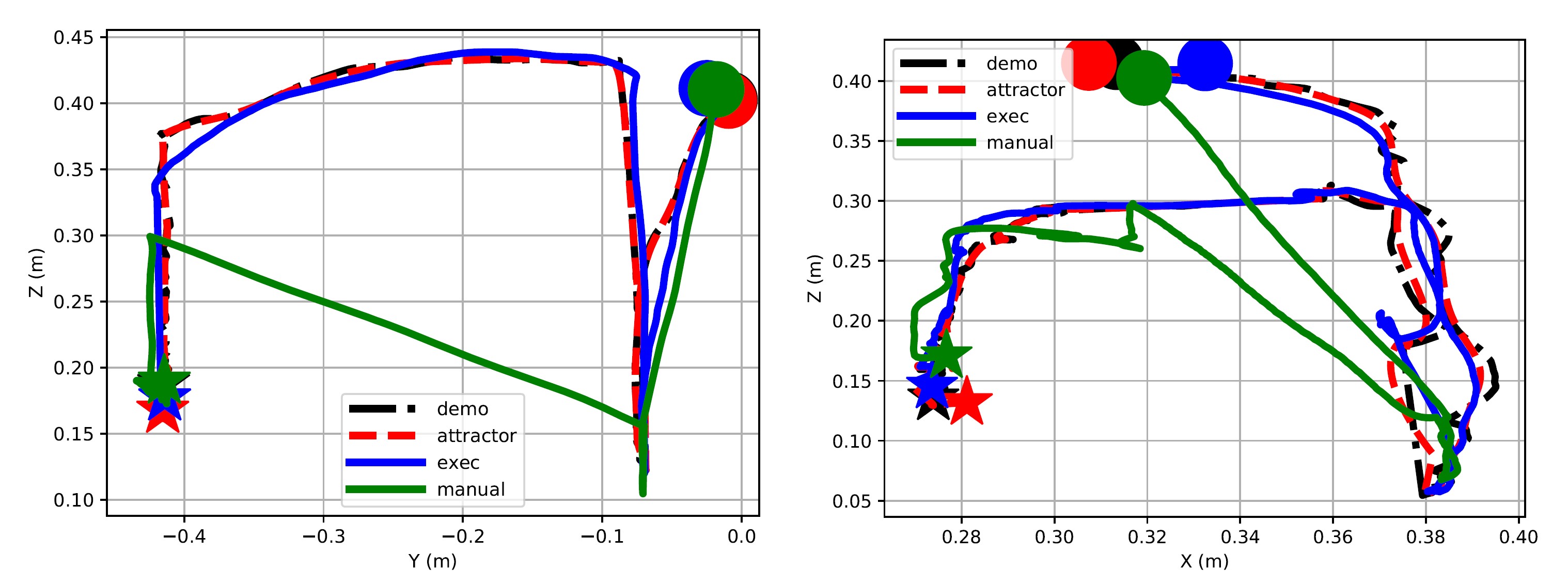}
  \caption{Comparison of executed trajectories via demo-replay, manual tunning and our method, for the Insert-Shaft skill (left) projected on $y-z$ plane and Slide-Peg skill (right) projected on $x-z$ plane.}
  \label{fig:compare}
  \vspace{-0.1cm}
\end{figure}

The resulting success rate is summarized in Table~\ref{table:results}. 
First, the success rate of demo-replay and pose-based methods  are quite low for all four skills, especially when delicate force interaction is required.
Often, via both methods, the skill execution simply reaching these key reference points without exerting the desired force, e.g.,  the robot only touches the pins without any pushing force during Press-PCB. 
Moreover, raw human demonstrations are quite \emph{shaky} in general as shown in Fig.~\ref{fig:compare}. 
Replaying such  small unnecessary movements are often harmful for the execution. 
Lastly, the manual tunning of both Insert-shaft and Slide-peg skills can lead to rather reliable execution. 
However, the time taken to program these skills is \emph{significantly} longer than the method proposed here (hours vs. minutes empirically). 
In addition,  the resulting trajectory often follows a zig-zag pattern with harsh transitions due to linear interpolation between manually-chosen waypoints, as shown in Fig.~\ref{fig:compare}. 

Note that our previous work~\cite{rozo2020learning} proposed a method to automatically choose the best grasping maneuver via the task parameterized skill model, given different object poses.  
However, due to the lack of reliable and accurate perception module, the poses of the work pieces are \emph{not} changed in this experiment, which remains part of our future work.

\section{Conclusion} \label{sec:conclusion}
This work extends the LfD framework to learn forceful manipulation skills from multi-modal demonstrations. 
The learned skill model consists of the attractor model and the stiffness model.
Furthermore, an online execution algorithm is proposed to adapt the skill execution to real-time observations.
Future work involves the combination with learning to mitigate the tracking errors during execution.

\bibliographystyle{IEEEtran}
\bibliography{references}

\end{document}